\documentclass[twoside,11pt]{scrartcl}

\usepackage{makeidx}
\usepackage{graphicx}
\usepackage{amssymb}
\usepackage{amsmath}
\usepackage{amscd}

\input xy
\xyoption{all}
\CompileMatrices

\newcommand{\qed}{\hspace*{\fill} $\blacksquare$}
\newcommand{\proof}{\noindent\textbf{Proof:}\quad}

\newtheorem{proposition}{Proposition}[section]

\newtheorem{example}{Example}[section]

\newcommand{\commentout}[1]{}

\newcommand{\N}{\mathbb{N}}                    
\newcommand{\R}{\mathbb{R}}                    
\newcommand{\E}{\mathbb{E}}                    

\newcommand{\abs}[1]{\mathop{\left\lvert #1 \right\rvert}} 
\newcommand{\args}[1]{\mathop{\left( #1 \right)}} 
\newcommand{\inner}[1]{\mathop{\left\langle #1 \right\rangle}}
\newcommand{\norm}[1]{\mathop{\left\lVert #1 \right\rVert}}
\newcommand{\cbrace}[1]{\mathop{\left\{ #1 \right\}}}
\newcommand{\bracket}[1]{\mathop{\left[ #1 \right]}}

\newcommand{\argsS}[2]{\mathop{\left( #1 \right)#2}} 

\newcommand{\normS}[2]{\mathop{\left\lVert #1 \right\rVert#2}}

\DeclareMathOperator{\conv}{con}               

\DeclareMathOperator{\id}{id}                  
\newcommand{\T}{\mathop{\mathsf{T}}}           

\renewcommand{\S}[1]{{\mathcal{#1}}}           
\def\vec#1{\mathchoice{\mbox{\boldmath$\displaystyle#1$}}
{\mbox{\boldmath$\textstyle#1$}}
{\mbox{\boldmath$\scriptstyle#1$}}
{\mbox{\boldmath$\scriptscriptstyle#1$}}}

{%
\end{array}\right.}

\newcounter{algorithm_counter}
{
\\[-1.5ex]
\rule{\textwidth}{\arrayrulewidth}
\end{sf}
\end{footnotesize}
\end{minipage}
\setlength{\parindent}{\parindent}
}

{
\\[-1.5ex]
\rule{\textwidth}{\arrayrulewidth}
\end{sf}
\end{footnotesize}
\end{minipage}
\setlength{\parindent}{\parindent}
}

\begin{document}

\title{Learning in Riemannian Orbifolds}

\author{Brijnesh J.~Jain and Klaus Obermayer\\
       Technische Universit\"at Berlin\\
       Berlin, Germany\\
       e-mail: brijnesh.jain@gmail.com}
            
\date{}
\maketitle

\begin{abstract} 
Learning in Riemannian orbifolds is motivated by existing machine learning algorithms that directly operate on finite combinatorial structures such as point patterns, trees, and graphs. These methods, however, lack statistical justification. This contribution derives consistency results for learning problems in structured domains and thereby generalizes learning in vector spaces and manifolds.
\end{abstract}

\section{Introduction}

Statistical data analysis and learning in Riemannian orbifolds is motivated by applications, where the data we want to learn on are naturally represented by finite combinatorial structures such as point patterns, trees, and graphs. Examples from structural pattern recognition that learn on structured data include estimating central points of a distribution on graphs such as the mean and median \cite{Ferrer:2007,Jain:2009d,Jain:2008,Jiang:2001}, central clustering of graphs \cite{Ferrer:2009b,Gold:1996b,Gunter:2002,Hlaoui:2006,Jain:2004a,Jain:2008,Schenker:2007}, learning graph quantization \cite{Jain:2009b}, and multilayer perceptrons for graphs \cite{Jain:2004b}. In retrospect, the structure space framework proposed by \cite{Jain:2009a} theoretically justifies the above approaches in the sense that they actually minimize an empirical risk function on structures. Since minimizing an empirical risk function is usually computationally intractable, the ultimate challenge consists in constructing efficient algorithms which are capable to return optimal or at least suboptimal solutions.

From the point of view of statistical pattern recognition, however, the ultimate goal is not to determine a good solution of an empirical risk function, but rather to discover the true but unknown structure of the data with respect to its distribution. According to this perspective, we may regard the solutions of empirical risk functions as estimators of the true but unknown population parameter. One gap between statistical and structural pattern recognition is the lack of consistency results of existing estimators for the population parameters. As a consequence most methods from structural pattern recognition that directly operate in the domain of graphs still have no  statistical justification.

The first contribution of this paper establishes sufficient conditions for consistency of estimators defined by empirical risk functions on attributed graphs. For this we regard graphs as points of some structure space \cite{Jain:2009a}. A structure space is the quotient of a Euclidean space by some permutation group. The benefit of the structure space framework  is that it provides enough mathematical structure for doing differential geometry and at the same time preserves the full relational information of the graphs. In comparison to \cite{Jain:2009a}, the innovations are as follows: First, we extend the more suitable concept of generalized differentiability in the sense of Norkin \cite{Norkin:1986}  to functions on graphs. Second, we prove the  stronger result that  the underlying empirical risk functions on graphs are generalized differentiable rather than locally Lipschitz. Third, equipped with these results, we apply a consistency theorem by Ermoliev and Norkin \cite{Ermoliev:1998} for generalized differentiable loss functions. Finally, using some examples, we show that standard methods from statistical pattern recognition can be generalized to consistent learning algorithms on graphs.

The second contribution shifts the terminology from structure spaces to the more general notion of orbifold. Informally, orbifolds are topological spaces locally modeled on quotients of manifolds by finite group actions. As such, structure spaces are the simplest examples of Riemannian orbifolds. Shifting the focus to orbifolds provides a new view on the problem with the following benefits: First, the notion of orbifold more strongly emphasizes the way we exploit differential geometric tools  for graphs, namely via charting and lifting as in Riemannian geometry. Second, using the notion of orbifold integrates the structure space framework into an established mathematical field providing access to useful concepts, results, and insights. Third, the notion of orbifold indicates how the theory can be generalized to structures that locally live in a quotient of a manifold by some finite group action. Fourth, since orbifolds generalize Euclidean spaces and manifolds, this framework not only establishes consistency for stochastic generalized gradient learning but also for standard stochastic gradient learning in Euclidean spaces (see \cite{Bottou:2003}) under the unifying umbrella of learning on Riemannian orbifolds.

 \section{The Problem of Learning on Graphs}

This section aims at outlining the problem of learning on structured data in order to motivate learning in Riemannian orbifolds. As an illustrative example, we consider the problem of estimating the mean of a distribution on attributed graphs.

\paragraph*{Attributed Graphs.}
We begin with describing the structures we want to learn on. Let $\S{A}$ be a set of \emph{attributes} and let $\varepsilon \in \S{A}$ be a distinguished element denoting the \emph{null} or \emph{void} element.  An  \emph{attributed graph} is a tuple $X = (V, \alpha)$ consisting of a finite nonempty set $V$ of \emph{vertices} and an \emph{attribute function} $\alpha: V \times V \rightarrow \S{A}$. Elements of the set $E = \cbrace{(i,j)\in V \times V \,:\, i \neq j \text{ and }\alpha(i,j) \neq \varepsilon}$ are the \emph{edges} of $X$. By $\S{G_A}$ we denote the set of all attributed graphs with attributes from $\S{A}$. The vertex set of an attributed graph $X$ is often referred to as  $V_X$ and its attribute function as $\alpha_X$.  

\paragraph*{Alignments.} 
Alignments serve to compare the common structure of two given graphs. An \emph{alignment} of a graph $X$ is a graph $X'$ with $V_X \subseteq V_{X'}$ and 
\[
\alpha_{X'}(i,j) = \begin{cases}
\alpha_X(i,j) & (i,j) \in V_X \times V_X\\
\varepsilon & \text{otherwise}
\end{cases} \qquad \forall \; i,j\in V_{X'}.
\]
Thus, we obtain an alignment of $X$ by adding isolated vertices with null-attribute. The set $V_{X'}^\varepsilon = V_{X'}\!\setminus\! V_X$ is the set of \emph{aligned vertices}. By $\S{A}(X)$ we denote the infinite set of all alignments of $X$. A \emph{pairwise alignment} of graphs $X$ and $Y$ is a triple $(\phi, X',Y')$ consisting of  alignments $X'\in \S{A}(X)$ and $Y'\in \S{A}(Y)$ together with a bijective mapping
\[
\phi : V_{X'} \rightarrow V_{Y'}, \quad i \mapsto i^{\phi}.
\]
A pairwise alignment $(\phi, X',Y')$ is \emph{minimal} if $\phi$ does not map aligned vertices onto each other, that is $\phi\args{V_{X'}^\varepsilon} \subseteq V_Y$.
By $\S{A}(X,Y)$ we denote the set of all minimal pairwise alignments between $X$ and $Y$. Note that $\S{A}(X, Y)$ is finite due to the minimality condition. Sometimes we briefly write $\phi$ instead of $(\phi, X',Y')$.

\paragraph*{Graph Edit Distance.}
Dissimilarity is a fundamental concept in machine learning. Here, we consider the graph edit distance, which is a common choice for measuring structural variation of two given graphs. Several distance measures reported in the structural pattern recognition literature can be derived as special cases of the graph edit distance function. Examples are  geometric graph distance functions \cite{Gold:1996a}  and distances based on the maximum common subgraph including graph and subgraph isomorphism \cite{Bunke:1997}.

To define the graph edit distance, we regard each minimal pairwise alignment $(\phi,X',Y') \in \S{A}(X,Y)$ as an \emph{edit path} with \emph{edit cost}
\[
d_\phi\args{X',Y'} = \sum_{i,j \in V_{X'}} d_{\S{A}}\args{\alpha_{X'}(i,j), \alpha_{Y'}(i^\phi, j^\phi)},
\]
where $d_\S{A}: \S{A} \times \S{A} \rightarrow \R_+$ is a distance function defined on the set $\S{A}$ of attributes. The edit cost $d_\phi$ can be decomposed into deletion cost $d_{A}(a, \varepsilon)$, insertion cost $d_{A}(\varepsilon, a')$, and substitution cost $d_{A}(a, a')$ of vertices and edges, where $a, a'\in \S{A}\setminus \cbrace{\varepsilon}$ are non-null attributes. Since $d_{\S{A}}$ is a distance function, we have $d_{\S{A}}(\varepsilon,\varepsilon) = 0$. This can only occur for pairs of non-edges by definition of minimal pairwise alignments and therefore can safely be ignored. Observe that deletion (insertion) of vertices also deletes (inserts) all edges the respective vertices are incident to. The \emph{graph edit distance} of $X$ and $Y$ is then defined as the edit path with minimal cost 
\[
d(X, Y) = \min\cbrace{d_\phi\args{X',Y'} \,:\, (\phi, X', Y') \in \S{A}(X,Y)}.
\]

\paragraph*{The Problem of Learning.}
Let $\args{\S{G_A}, d}$ be a graph distance space. As an illustrative example, consider the expected risk
\[
R(W) = \frac{1}{2}\int_{\S{G_A}} d\!\argsS{X, W}{^2} dP_{\S{G_A}}(X),
\]
where $W\in \S{W} \subseteq \S{G_A}$ is the optimization variable and $X\in \S{G_A}$ is a random variable with probability distribution $P_{\S{G_A}}$. Since the distribution on the set $\S{G_A}$ of graphs is usually unknown, the goal of learning is to minimize the risk $R(W)$ on the basis of empirical data.

To point out the problems of learning in the domain of graphs, we consider the counterpart of minimizing the risk $R(W)$ in a Euclidean vector space $\S{X}$. The goal is to minimize the expected risk
\[
R(\vec{w}) = \frac{1}{2}\int_{\S{X}} \normS{\vec{x}-\vec{w}}{^2} dP_{\S{X}}(\vec{x}),
\]
based on independent and identically distributed random points $\vec{x}_1, \ldots, \vec{x}_N \in \S{X}$, where $P_{\S{X}}$ is a probability measure on $\S{X}$. Since the loss function $\normS{\vec{x}-\vec{w}}{^2}$ is continuously differentiable, the interchange of integral and gradient is valid, that is
\[
\nabla R(\vec{w}) = -\int_{\S{X}} (\vec{x}-\vec{w}) dP_{\S{X}}(\vec{x}).
\]
We can minimize the risk $R(\vec{w})$ using the following stochastic gradient method
\[
\vec{w}_{t+1} = \vec{w}_t + \frac{1}{t+1}\args{\vec{x}_t-\vec{w}_t},
\]
where $\vec{w}_1 = \vec{x}_1$ and $t \geq 1$. The elements $\vec{w}_t$ of the sequence $(\vec{w}_t)_{t\geq 0}$ are sample means
\[
\vec{w}_t = \frac{1}{t}\sum_{i=1}^t \vec{x}_t.
\]
It is well-known that the sample mean is a consistent estimator of the population mean $\vec{\mu}$, which in turn is the unique global minimizer of the expected risk $R(\vec{w})$.

After this short digression in vector spaces, let us return to the problem of minimizing the expected risk $R(W)$ in graph spaces. As opposed to vector spaces, the following factors complicate learning on graphs in a statistically consistent way:  (i) the graph edit distance $d(X, Y)$ is in general not-differentiable; and (ii) neither a well-defined addition on graphs nor the notion of derivative for functions on graphs is known.

We therefore address the following questions: (i) How can we extend gradient-based learning problems from Euclidean spaces to $\S{G_A}$? (ii) How can we minimize the expected risk of a learning problem with structured input- and/or output-space $\S{G_A}$ in a statistically consistent way?

The ansatz to answer both questions is to identify graphs as points of a Riemannian orbifold and to extend the concept of generalized differentiability in the sense of Norkin \cite{Norkin:1986} in order to apply methods from stochastic optimization for non-differentiable and non-convex loss functions. 

\section{Riemannian Orbifolds}

This section introduces Riemannian orbifolds. To keep the treatment technically as uncluttered as possible, we assume that $\S{X} = \R^{n}$ is the $n$-dimensional Euclidean space, and $\Gamma$ is a permutation group acting on $\S{X}$. In doing so, we can refer to \cite{Jain:2009a} for proofs of statements and claims made in this section. In a more general setting, however, $\S{X}$ can also be a Riemannian manifold. In this case, we refer to \cite{Borzellino:1992} for more details. 

\subsection{Riemannian Orbifolds}

The binary operation
\[
\cdot: \Gamma \times \S{X} \rightarrow \S{X}, \quad (\gamma, \vec{x}) \mapsto \gamma(\vec{x})
\]
is a group action of $\Gamma$ on $\S{X}$. For $\vec{x} \in \S{X}$, the \emph{orbit} of $\vec{x}$ is the set defined by $\bracket{\vec{x}} = \cbrace{\gamma(\vec{x}) \,:\, \gamma \in \Gamma}$. The quotient set $\S{X}_\Gamma = \S{X}/\Gamma = \cbrace{\bracket{\vec{x}} \,:\, \vec{x} \in \S{X}}$ consisting of all all orbits  carries the structure of a \emph{Riemannian orbifold}. Its \emph{orbifold chart} is the surjective continuous mapping 
\[
\pi: \S{X} \rightarrow \S{X}_\Gamma, \quad \vec{x} \mapsto \bracket{\vec{x}}
\]
that projects each point $\vec{x}$ to its orbit $\bracket{\vec{x}}$. With $\Gamma =\cbrace{\id}$ being the trivial permutation group, $\S{X}$ is also an orbifold. Hence, orbifolds generalize the notion of Euclidean space and manifold. 

In the following, an orbifold is a triple $\S{Q} = \args{\S{X}, \Gamma, \pi}$ consisting of a Euclidean space $\S{X}$, a permutation group $\Gamma$ acting on $\S{X}$ and its orbifold chart $\pi$. We call the elements of $\S{X}_\Gamma$ \emph{structures}, since they represent combinatorial structures such as graphs. We use capital letters $X, Y, Z, \ldots$ to denote structures from $\S{X}_\Gamma$ and write $\vec{x} \in X$ if $\pi(\vec{x}) = X$. Each vector $\vec{x} \in X$ is a \emph{vector representation} of structure $X$ and the set $\S{X}$  of all vector representation is the \emph{representation space} of $\S{X}_\Gamma$.

\subsection{The Riemannian Orbifold of Graphs}

Riemannian orbifolds of attributed graphs arise by considering equivalence classes of matrices representing the same graph. To identify graphs with points from some orbifold, some technical assumptions  to simplify the mathematical treatment are necessary. For this, let $\args{\S{G_A}, d}$ be a graph distance space with graph edit distance $d(\cdot|\cdot)$. Then we make the following assumptions:

A1. There is a feature map $\Phi:\S{A} \rightarrow \S{H}$ of the attributes into some finite dimensional Euclidean feature space $\S{H}$ and a distance function $d_{\S{H}}: \S{H}\times \S{H} \rightarrow \R_+$ such that $\Phi(\varepsilon) = \vec{0} \in \S{H}$ and
\[
d_{\S{A}}(a,a') = d_{\S{H}}(\Phi(a), \Phi(a')) \quad \forall\;a, a'\in \S{A}.
\]

A2. All graphs are finite of bounded order $n$, where $n$ is a sufficiently large number. A graph $X$ of order less than $n$, say $m < n$, is aligned to graph $X'$ of order $n$ by inserting $p = n-m$ isolated vertices with null attribute $\varepsilon$. 

Let us consider the above assumptions in more detail. Both conditions do not effect the graph edit distance, provided an appropriate feature map for the attributes can be found. Restricting to finite dimensional Euclidean feature spaces $\S{H}$ is necessary for deriving consistency results and for applying methods from stochastic optimization. Limiting the maximum size of the graphs to some arbitrarily large number $n$ and aligning smaller graphs to graphs of oder $n$ are purely technical assumptions to simplify mathematics. For machine learning problems, this limitation should have no practical impact, because neither the bound $n$ needs to be specified explicitly nor an extension of all graphs to an identical order needs to be performed. When applying the theory, all we actually require is that the order of the graphs is bounded.

With both assumptions in mind, we construct the Riemannian orbifold of attributed graphs. Let $\S{X} = \S{H}^{n \times n}$ be the set of all $(n \times n)$-matrices with elements from feature space $\S{H}$. A graph $X$ is completely specified by a \emph{representation matrix} $\vec{X} = \args{\vec{x}_{ij}}$ from $\S{X}$ with elements 
\[
\vec{x}_{ij} = \begin{cases}
\phi\args{\alpha_X(i, j)} & i = j \mbox{ or } (i,j) \in E\\
\vec{0} & \mbox{otherwise}
\end{cases}
\]
for all $i, j \in V_X$. The form of a representation matrix $\vec{X}$ of $X$ is generally not unique and depends on how the vertices are arranged in the diagonal of $\vec{X}$.

Now suppose that $\Pi^n$ be the set of all $(n\times n)$-permutation matrices. For each $\vec{P} \in \Pi^n$ we define a mapping
\[
\gamma_{\vec{P}} : \S{X} \rightarrow \S{X}, \quad \vec{X} \mapsto \vec{P}^{\T} \vec{X}\vec{P}.
\]
Then $\Gamma = \cbrace{\gamma_{\vec{P}}\,:\, \vec{P} \in \Pi^n}$ is a permutation group acting on $\S{X}$. Regarding an arbitrary matrix $\vec{X}$ as a representation of some graph $X$,  then the orbit $\bracket{\vec{X}}$ consists of all possible matrices that can represent $X$. By identifying the orbits of $\S{X}_\Gamma$ with attributed graphs,  the set $\S{G_A}$ of attributed graphs of bounded order $n$ is a Riemannian orbifold.

\subsection{Metric Structures}
Let  $\S{Q} = \args{\S{X}, \Gamma, \pi}$ be an orbifold. We derive an intrinsic  metric that enables us to do Riemannian geometry. Note that in the case of graph orbifolds, the intrinsic metric is a special graph edit distance based on a generalization of the concept of maximum common subgraph. This graph metric occurs in various different guises as a common choice of proximity measure \cite{Almohamad:1993,Caetano:2007,Cour:2007,Gold:1996a,Umeyama:1988,Wyk:2002}.

Any inner product $\inner{\cdot, \cdot}$ on $\S{X}$ gives rise to a maximizer $k: \S{X}_\Gamma \times \S{X}_\Gamma \rightarrow \R$ of the form
\[
k(X, Y) = \max \cbrace{\inner{\vec{x}, \vec{y}} \,:\, \vec{x} \in X, \vec{y} \in Y}. 
\]
We call the kernel function $k(\cdot | \cdot)$ \emph{optimal alignment kernel}, induced by $\inner{\cdot, \cdot}$. Note that the maximizer of a set of positive definite kernels is an indefinite kernel in general. Since $\Gamma$ is a group, we find that 
\[
k(X, Y) = \max \cbrace{\inner{\vec{x}, \vec{y}} \,:\, \vec{x} \in X},
\]
where $\vec{y}$ is an arbitrary but fixed vector representation of $Y$. 

\begin{example}
Suppose that $X$ and $Y$ are attributed graphs where edges have attribute $1$ and vertices have attribute $0$. The optimal alignment kernel $k\args{X, Y}$ induced by the standard inner product of $\S{X}$ is the number of edges of a maximum common subgraph of $X$ and $Y$. 
\end{example}

Suppose that $X \in \S{X}_\Gamma$. Since $k(X, X) = \inner{\vec{x}, \vec{x}}$ for all $\vec{x} \in X$, we can define the \emph{length} of $X$  by 
\[
l(X) = \sqrt{k(X, X)}.
\]
Since the Cauchy-Schwarz inequality $\abs{k(X, Y)} \leq l(X)\cdot l(Y)$ is valid, the geometric interpretation of $k(\cdot|\cdot)$ is that it computes the cosine of a well-defined angle between $X$ and $X'$ provided both are normalized.

Likewise, $k(\cdot|\cdot)$ gives rise to a distance function defined by
\[
d(X, Y) = \sqrt{l(X)^2 -2 k(X, Y) +\, l(Y)^2}.
\]
From the definition of $k(\cdot|\cdot)$ follows that $d$ is a metric. In addition, we have
\begin{align}\label{eq:mcs-metric-2}
d(X, Y) =  \min \cbrace{\norm{\vec{x} - \vec{y}} \,:\, \vec{x} \in X, \vec{y} \in Y}, 
\end{align}
where $\norm{\cdot}$ denotes the Euclidean norm induced by the inner product $\inner{\cdot, \cdot}$ of the Euclidean space $\S{X}$. 

Equation (\ref{eq:mcs-metric-2}) states that $d\args{\cdot|\cdot}$ is the length of a minimizing geodesic of $X$ and $Y$ and therefore an intrinsic metric, because it coincides with the infimum of the length of all admissible curves from $X$ to $Y$.   In addition, we find that the topology of $\S{X}_\Gamma$ induced by the metric $d$ coincides with the quotient topology induced by the topology of the Euclidean space $\S{X}$. 

\subsection{Orbifold Mappings}

This section introduces  mappings between orbifolds and investigates local analytical concepts of orbifold functions. We assume that $\S{Q} = \args{\S{X}, \Gamma, \pi}$ and $\S{Q}' = \args{\S{X}', \Gamma', \pi'}$ are orbifolds.

\paragraph*{Mappings.}
An \emph{orbifold mapping} between $\S{Q}$ and $\S{Q'}$ is a mapping $f:\S{X}_{\Gamma} \rightarrow \S{X}'_{\Gamma'}$ between their underlying spaces. The \emph{lift} of $f$ is a mapping $\tilde{f}: \S{X} \rightarrow \S{X}'$ between their representation spaces such that $f\circ\pi =\pi'\circ\tilde{f}$.  Since $\R$ is an orbifold of the form $\S{Q}_\R = \args{\R, \cbrace{\id}, \id_\R}$, we can define an  \emph{orbifold function}  between $\S{Q}$ and $\S{Q}_\R$ as a function $f:\S{X}_{\Gamma} \rightarrow \R$. The lift of $f$ is a function $\tilde{f}: \S{X} \rightarrow \R$ satisfying $\tilde{f} = f \circ \pi$. The lift $\tilde{f}$ is invariant under group actions of $\Gamma$, that is $\tilde{f}(\vec{x}) = \tilde{f}\args{\gamma(\vec{x})}$ for all $\gamma \in \Gamma$.

We say, an orbifold function $f:\S{X}_{\Gamma} \rightarrow \R$ is continuous (locally Lipschitz, differentiable) at $X \in \S{X}_\Gamma$ if its lift $\tilde{f}$ is continuous (locally Lipschitz, differentiable) at some vector representation $\vec{x} \in X$. The definition is independent of the choice of the vector representation that projects to $X$. 

\paragraph*{Gradients.}
Suppose that $f: \S{X}_{\Gamma} \rightarrow \R$ is differentiable at $X \in \S{X}_\Gamma$. Then its lift $\tilde{f}:\S{X} \rightarrow \R$ is differentiable at all vector representations that project to $X$. The \emph{gradient} $\nabla f(X)$ of $f$ at $X$ is defined by the projection
\[
\nabla f(X) = \pi(\nabla \tilde{f}(\vec{x}))
\]
of the gradient $\nabla \tilde{f}(\vec{x})$ of $\tilde{f}$ at a vector representation $\vec{x} \in X$. This definition is independent of the choice of the vector representation. We have
\[
\nabla \tilde{f}(\gamma(\vec{x})) = \gamma(\nabla\tilde{f}(\vec{x}))
\]
for all $\gamma \in \Gamma$. This implies that the gradients of $\tilde{f}$ at $\vec{x}$ and $\gamma(\vec{x})$ are vector representations of the same structure, namely the gradient $\nabla f(X)$ of the orbifold function $f$ at $X$. Thus, the gradient of $f$ at $X$ is a well-defined structure pointing to the direction of steepest ascent.

\section{Generalized Gradients}

This section extends the concept of generalized differentiability in the sense of Norkin \cite{Norkin:1986} to  orbifold functions. We begin with introducing generalized differentiable functions. Let $\S{X} = \R^n$ be a finite-dimensional Euclidean space. A function $f: \S{X} \rightarrow \R$ is \emph{generalized differentiable} at $\vec{x}\in\S{X}$ if there is a multi-valued map $\partial f: \S{X} \rightarrow 2^{\S{X}}$ in a neighborhood of $\vec{x}$ such that 
\begin{enumerate}
\item $\partial f(\vec{x})$ is a convex and compact set;
\item $\partial f(\vec{x})$ is upper semicontinuous at $\vec{x}$, that is, if $\vec{y}_i \to \vec{x}$ and $\vec{g}_i \in \partial f(\vec{y}_i)$ for each $i\in \N$, then each accumulation point $\vec{g}$ of $(\vec{g}_i)$ is in $\partial f(\vec{x})$;
\item for each $\vec{y} \in \S{X}$ and any $\vec{g} \in \partial f(\vec{y})$ holds $f(\vec{y}) = f(\vec{x}) + \inner{\vec{g}, \vec{y}-\vec{x}} + o\args{\vec{x}, \vec{y}, \vec{g}}$,
where the remainder $o\args{\vec{x}, \vec{y}, \vec{g}}$ satisfies the condition
\[
\lim_{i\to \infty} \frac{\abs{o\args{\vec{x}, \vec{y}_i, \vec{g}_i}}}{\norm{\vec{y}_i -\vec{x}}} = 0
\]
for all sequences $\vec{y}_i \to \vec{y}$ and $\vec{g}_i \in \partial f\args{\vec{y}_i}$. 
\end{enumerate}
We call $f$ \emph{generalized differentiable} if it is generalized differentiable at each point  $\vec{x}\in\S{X}$. The set $\partial f(\vec{x})$ is the \emph{subdifferential} of $f$ at $\vec{x}$ and its elements are called \emph{generalized gradients}. 

Generalized differentiable functions have the following properties \cite{Norkin:1986}:
\begin{enumerate}
\item Generalized differentiable functions are locally Lipschitz and therefore continuous and differentiable almost everywhere.
\item Continuously differentiable, convex, and concave functions are generalized differentiable.
\item Suppose that $f_1, \ldots, f_n:\S{X} \rightarrow \R$ are generalized differentiable at $\vec{x}\in \S{X}$. Then 
\begin{align*}
f_*(\vec{x}) = \min(f_1(\vec{x}), \ldots, f_m(\vec{x})) \quad \mbox{and} \quad
f^*(\vec{x}) = \max(f_1(\vec{x}), \ldots, f_m(\vec{x}))
\end{align*}
are generalized differentiable at $\vec{x}\in \S{X}$.
\item Suppose that $f_1, \ldots, f_m:\S{X} \rightarrow \R$ are generalized differentiable functions at $\vec{x}\in \S{X}$ and $f_0:\R^m \rightarrow \R$ is generalized differentiable at $\vec{y} = \args{f_1(\vec{x}), \ldots, f_m(\vec{x})} \in\R^m$. Then $f(\vec{x}) = f_0(f_1(\vec{x}),\ldots, f_m(\vec{x}))$ is generalized differentiable at $\vec{x} \in \S{X}$. The subdifferential of $f$ at $\vec{x}$ is of the form
\begin{align*}
\partial f(\vec{x}) = \conv \Big\{\vec{g} \in \S{X} &:\, \vec{g} = \big[\vec{g}_1 \vec{g}_2\ldots \vec{g}_m\big]\vec{g}_0, \vec{g}_0 \in \partial f_0(\vec{y}), \vec{g}_i \in \partial f_i(\vec{x}), 1 \leq i \leq m \Big\}.
\end{align*}
where $\bracket{\vec{g}_1 \vec{g}_2\ldots \vec{g}_m}$ is a ($N\times m$)-matrix.
\item Suppose that $F(\vec{x}) = \E_{\vec{z}}\bracket{f(\vec{x}, \vec{z})}$, where $f(\cdot, \vec{z})$ is generalized differentiable.  Then $F$ is generalized differentiable and its subdifferential at $\vec{x} \in \S{X}$ is of the form $\partial F(\vec{x}) = \E_{\vec{z}}\bracket{\partial f(\vec{x},\vec{z})}$.
\end{enumerate}

Now suppose that $f: \S{X}_{\Gamma} \rightarrow \R$ is an orbifold function. We say $f$ is generalized differentiable at $X \in \S{X}_\Gamma$, if its lift $\tilde{f}:\S{X} \rightarrow \R$ is generalized differentiable at all vector representations that project to $X$. The \emph{subdifferential} $\partial f(X)$ of $f$ at $X$ is defined by the projection
\[
\partial f(X) = \pi(\partial \tilde{f}(\vec{x}))
\]
of the subdifferential $\partial \tilde{f}(\vec{x})$ of $\tilde{f}$ at a vector representation $\vec{x} \in X$. This definition is independent of the choice of the vector representation. We have 
\[
\partial \tilde{f}(\gamma(\vec{x})) = \gamma(\partial\tilde{f}(\vec{x}))
\]
for all $\gamma \in \Gamma$. This implies that the subdifferentials $\partial\tilde{f}(\vec{x}) \subseteq \S{X}$ and $\partial\tilde{f}(\gamma(\vec{x})) \subseteq \S{X}$ are subsets that project to the same subset of $\S{X}_\Gamma$, namely the subdifferential $\partial f(X)$. Proposition \ref{prop:gendiff} summarizes and proves the statements.

\begin{proposition}\label{prop:gendiff}
Let $f:\S{X}_{\Gamma} \rightarrow \R$ be an orbifold function. Suppose that its lift $\tilde{f}: \S{X} \rightarrow \R$  is generalized differentiable at a vector representation $\vec{x}$ that projects to $X \in \S{X}_\Gamma$. Then $\tilde{f}$ is generalized differentiable at $\gamma(\vec{x})$ for all $\gamma \in \Gamma$ and
\[
\partial \tilde{f}(\gamma(\vec{x})) = \gamma\args{\partial\tilde{f}(\vec{x})}.
\]
is a subdifferential of $\tilde{f}$ at $\gamma(\vec{x})$ for all $\gamma \in \Gamma$.
\end{proposition}

\proof Since $\tilde{f}$ is generalized differentiable at $\vec{x}$, there is a multi-valued mapping $\partial\tilde{f}: \S{U}_\delta(\vec{x})\rightarrow 2^{\S{X}}$ defined on some neighborhood $\S{U}_\delta(\vec{x})$. Let $\gamma \in \Gamma$ be an arbitrary permutation and $\vec{x}' = \gamma(\vec{x})$. Then 
\[
\partial\tilde{f}: \S{U}_\delta(\vec{x}')\rightarrow 2^{\S{X}}, \quad \vec{y}' = \gamma(\vec{y}) \mapsto \gamma\args{\partial \tilde{f}(\vec{y})}
\]
is a multi-valued mapping in a neighborhood of $\vec{x}'$. Since $\gamma$ is a homeomorphic linear map, we find that $\gamma(\partial\tilde{f}(\vec{x})) = \partial \tilde{f}(\vec{x}')$ is a convex and compact set. Next we show that $\tilde{f}$ is upper semicontinuous at $\vec{x}'$.  Suppose that $\vec{y}'_i \to \vec{x}'$, $\vec{g}'_i \in \tilde{f}_c(\vec{y}'_i)$ for each $i\in \N$, and $\vec{g}'$ is an accumulation point of $(\vec{g}'_i)_{i \in \N}$.  Then there is a $i_0 \in \N$ such that $\vec{y}'_i \in \S{U}_\delta(\vec{x}')$ for all $i \geq i_0$. From 
\[
\S{U}_\delta(\vec{x}') = \S{U}_\delta(\gamma(\vec{x})) = \gamma\args{\S{U}_\delta(\vec{x})}
\]
follows that there are vector representations $\vec{y}_i \in \S{U}_\delta(\vec{x})$ with $\gamma(\vec{y}_i) = \vec{y}_i'$ for each $i \geq i_0$. From continuity of $\gamma^{-1}$ follows that $\vec{y}_i \to \vec{x}$. By construction of $\partial \tilde{f}$ follows that
\[
\vec{g}_i' \in \partial\tilde{f}\args{\vec{y}_i'} = \partial\tilde{f}\args{\gamma\args{\vec{y}_i}} = \gamma\args{\partial\tilde{f}\args{\vec{y}_i}}
\]
for each $i\geq i_0$. Hence, there are vector representations $\vec{g}_i \in \partial\tilde{f}(\vec{y}_i)$ with $\gamma(\vec{g}_i) = \vec{g}_i'$ for each $i \geq i_0$. Since $\tilde{f}$ is upper semicontinuous at $\vec{x}$, we find that $\vec{g} \in \partial\tilde{f}(\vec{x})$. Again by construction of $\partial \tilde{f}$ follows that  
\[
\vec{g}' = \gamma(\vec{g}) \in \gamma\args{\partial\tilde{f}(\vec{x})} = \partial\tilde{f}\args{\gamma(\vec{x})} = \partial \tilde{f}(\vec{x}').
\]
This proves upper semicontinuity of $\partial \tilde{f}$ at all vector representations projecting to $X = \pi(\vec{x})$. Finally, we prove that $\tilde{f}$ satisfies the subderivative property at $\vec{x}'$. Suppose that $\vec{y}', \vec{y} \in \S{X}$ with $\vec{y}' = \gamma(\vec{y})$. By $\Gamma$-invariance of $\tilde{f}$, we have $\tilde{f}(\vec{y}') = \tilde{f}(\vec{y})$. Since $\tilde{f}$ is generalized differentiable at $\vec{x}$, we find a $\vec{g} \in \partial \tilde{f}(\vec{y})$ such that
\[
\tilde{f}(\vec{y}') = \tilde{f}(\vec{y}) = \tilde{f}(\vec{x}) + \inner{\vec{g}, \vec{y}-\vec{x}} +\, o(\vec{x}, \vec{y}, \vec{g})
\]
with $o(\vec{x}, \vec{y}, \vec{g})$ tending faster to zero than $\norm{\vec{y}-\vec{x}}$. Let $\vec{g}' = \gamma(\vec{g})$. Exploiting $\Gamma$-invariance of $\tilde{f}$ as well as isometry and linearity of $\gamma$ yields
\begin{align*}
\tilde{f}(\vec{y}') &= \tilde{f}(\gamma(\vec{x})) + \inner{\gamma(\vec{g}),\gamma(\vec{y}-\vec{x})} +\, o(\vec{x}, \vec{y}, \vec{g})\\
&= \tilde{f}(\vec{x}') + \inner{\vec{g}',\vec{y}'-\vec{x}'} +\, o(\vec{x}, \vec{y}, \vec{g}).
\end{align*}
We define $o'(\vec{x}', \vec{y}', \vec{g}') = o\circ \gamma^{-1}(\vec{x}', \vec{y}', \vec{g}') = o(\vec{x}, \vec{y}, \vec{g})$ showing that $o'$ tends faster to zero than $\norm{\vec{y}' -\vec{x}}$. This proves the subderivative property of $\tilde{f}$  at all vector representations projecting to $X = \pi(\vec{x})$.  Putting all results together yields that $\tilde{f}$ is generalized differentiable at $\gamma(\vec{x})$ for all $\gamma \in \Gamma$.
\qed

\begin{example}\label{ex:gd-distortion1}
Let $(\S{G_A}, d)$ be a graph space, where $d$ is a graph edit distance.We can identify $\S{G_A}$ with a Riemannian orbifold $\S{Q} = (\S{X}, \Gamma, \pi)$ and the graph edit distance $d\args{\cdot|\cdot}$ with a distance function defined on $\S{X}_\Gamma$. Suppose that the edit costs $d_\phi\args{\cdot|\cdot}$ of all edit paths are generalized differentiable. Then the distance  $d\args{\cdot|\cdot}$ is generalized differentiable.
\end{example}

\begin{example}\label{ex:gd-distortion2}
Let $\S{Q}$ be a graph orbifold. Then the optimal assignment kernel $k\args{\cdot|\cdot}$, the intrinsic metric  $d\args{\cdot|\cdot}$, and the squared metric $d\args{\cdot|\cdot}^{2}$ are generalized differentiable. 
\end{example}

\section{Stochastic Optimization}
\newcommand{\sfH}{\mathsf{H}}

We assume that $\S{Q_W} = (\S{W}, \mathsf{H}, \rho)$ and $\S{Q_Z} = (\S{Z}, \Gamma, \pi)$ are Riemannian orbifolds and $\Omega \subseteq \S{W}_{\sfH}$ is some (sufficiently large) bounded convex constraint set. Learning is formulated as a stochastic optimization problem of the form
\begin{align}
\label{eq:erm-obj}
\min \; &R(W) = \E\bracket{L(Z,W)} = \int_{\S{Z}_\Gamma} L(Z,W) dP_{\Gamma}(Z)\\
\label{eq:erm-con}
\mbox{s.t.} \; & W \in \Omega,
\end{align}
where $R(W)$ is the \emph{expected risk function}, $W \in \Omega$ is the optimization variable, and $Z\in \S{Z}_\Gamma$ is a random variable with probability measure $P_{\Gamma}$. The \emph{loss function} $L:\S{Z}_\Gamma\times \Omega \rightarrow \R$ measures the performance of the learning system with parameter $W$ given an observable event $Z$. We assume that the loss $L(Z,W)$ is generalized differentiable in $W$ and integrable in $Z$. The expectation $\E$ is taken with respect to some probability space $\args{\S{Z}_\Gamma, \Sigma_{\Gamma}, P_{\Gamma}}$. 

Since the distribution $P_{\S{Z}}$ of the  observable events $Z\in \S{Z}$ is usually unknown, the expected risk function $R(W)$ can neither be computed nor be minimized directly.  In addition,  the loss function $L(W,Z)$ is neither convex nor differentiable. The field of stochastic approximation provides methods to minimize $R(W)$ that are consistent under very general conditions. 

Since  the interchange of integral and generalized gradient is valid, that is $\partial_W R(W) = \E\bracket{\partial_W L(Z,W)}$ under mild assumptions \cite{Ermoliev:1998,Norkin:1986}, we can minimize the expected risk $R(W)$ according to the following \emph{stochastic generalized gradient} (SGG) method:
\begin{align*}
W_{t+1} &= \Pi_\Omega\args{W_t - \eta_t  S_t}, \qquad t \geq 0,
\end{align*}
where $W_0 \in \Omega$ and $\Pi_\Omega$ is a projection operator on $\Omega$. The random structures $S_t$ are \emph{stochastic generalized gradients}, i.e.\ random variables defined on the probability space $\argsS{\S{Z}_\Gamma, \Sigma_{\Gamma}, P_{\Gamma}}{^\infty}$ such that 
\begin{align}\label{eq:A2}
\E\bracket{S_t \,|\, W_0, \ldots, W_t} \in \partial_W R\args{W}.
\end{align}
We can take $S_t = g(Z_t, W_t)$ with iid $\argsS{Z_t}{_{t\geq 0}}$ and some single valued selection $g(Z, W) \in \partial_{W} L(Z, W)$, measurable in $(Z,W)$.
We consider the following conditions for almost sure convergence of the SSG method:
\begin{description}
\item[A1] The sequence $(\eta_t)_{t \geq 0}$ of step sizes satisfies 
\[
\eta_t > 0, \; \lim_{t \to \infty} \eta_t = 0, \; \sum_{t=1}^{\infty} \eta_t = \infty, \; \sum_{t=1}^{\infty} \eta_t^2 < \infty.
\]
\item[A2] The sequence $\args{S_t}_{t \geq 0}$ satisfies (\ref{eq:A2}).
\item[A3]  We have $\E\bracket{\normS{S_t}{^2}} < +\infty$.
\end{description}
Then by Ermoliev and Norkin's Theorem \cite{Ermoliev:1998}, the SGG method is consistent in the sense that the sequence $\argsS{W_t}{_{t\geq 0}}$ converges almost surely to points satisfying necessary extremum conditions 
\[
\Omega^* = \cbrace{W \in\Omega \,:\; 0 \in \partial_{W} R(W) +  \S{N}_{\Omega}(W)},
\]
where $\S{N}_{\Omega}(W)$ is a normal cone to the constraint set $\Omega$ at $W \in \Omega$. Besides the sequence $\argsS{R(W_t)}{_{t\geq 0}}$ converges almost surely and $\lim_t R(W_t) \in R(\Omega^*)$.

Since orbifolds generalize Euclidean spaces and manifolds the consistency theorem is also valid for standard machine learning algorithms in Euclidean spaces with differentiable cost function (e.g multi-layer perceptron) and non-differentiable cost function (e.g. online k-means) \cite{Bottou:2003}.  

\section{Examples}

This section extends some typical examples of statistical data analysis and learning problems from vector spaces to structured domains. We assume that $\S{Q} = \args{\S{X}, \Gamma, \pi}$ is a Riemannian orbifold with optimal alignment kernel $k(\cdot|\cdot)$.

\paragraph*{Orbifold-Adaline.}
Orbifold adaline generalizes the \emph{adaline} proposed by \cite{Widrow:1960}.

Let $\S{W} = \S{X}_\Gamma \times R$ be the parameter space and let $\S{Z} = \S{X}_\Gamma \times \cbrace{\pm 1}$ be the space of observable data. The parameter space $\S{W}$ consists of augmented parameter structures $W' = (W, b)$, where $W \in \S{X}_\Gamma$ is the weight structure and $b \in \R$ is the bias. The observable data $Z = (X, y)$ from $\S{Z}$ consists of input structures $X \in \S{X}_\Gamma$ together with their labels $y \in \cbrace{\pm 1}$.  

The loss function of the orbifold-Adaline is of the form
\[
L_{ada}(Z, W') = \big(y - (k(X,W) + b)\big)^2.
\]
Since $k(\cdot|\cdot)$ is generalized differentiable, so is $L_{ada}(Z,W)$. Lifting the loss $L_{ada}$ to the Euclidean space gives
\[
\hat{L}_{ada}\args{\vec{z}, \vec{w}'} = \args{y - \max\cbrace{\inner{\vec{x}', \vec{w}}\,:\, \vec{x}'\in X} - \, b}{^2},
\]
where $\vec{z} = (\vec{x}, y) \in \S{Z}$ and $\vec{w}' = (\vec{w}, b) \in \S{W}$ with vector representations $\vec{x}$ and $\vec{y}$ that project to structures $X\in\S{X}_\Gamma$ and $W\in \S{X}_\Gamma$, respectively. The update rule is given by
\begin{align*}
\vec{w}_{t+1} &= \vec{w}_t - \eta_t\args{y_t - \inner{\vec{x}_t^*, \vec{w}_t}\vec{x}_t^*} \\
b_{t+1} &= b_t - \eta_t \args{y_t - b_t},
\end{align*}
where $(\vec{x}_t^*, \vec{w}_t)$ is an optimal alignment.

\paragraph*{Learning Orbifold Maps.}
This example presents a generic formulation of learning functional relationships between orbifolds in a supervised manner. Since orbifolds generalize Euclidean spaces, this setting  covers various types of functional relationships that can be learned. Non-standard examples include multi-layer perceptrons for adaptive processing of graphs \cite{Jain:2004b} and learning to predict structured data \cite{Bakir:2007}.

Let $\S{Q_W} = \args{\S{W}, \Omega, \psi}$,  $\S{Q_X} = \args{\S{X}, \Gamma, \pi}$, and $\S{Q_Y} = \args{\S{Y}, \Lambda, \phi}$ be Riemannian orbifolds. The parameter space is represented by orbifold $\S{Q_W}$ and the space of observable data by the orbifold $\S{Q_Z} = \S{Q_X} \times \S{Q_Y}$. Suppose that $\S{F}$ is a class of generalized differentiable orbifold mappings of the form 
\[
f:\S{X}_\Gamma \times \S{W}_\Omega \rightarrow \S{Y}_\Lambda.
\]
The mean-squared-error loss function is defined by
\[
L_{mse}(Z, W) = \frac{1}{2}\argsS{Y - f(X,W)}{^2}.
\]
Lifting this loss function yields
\[
\hat{L}_{mse}(\vec{z}, \vec{w}) = \frac{1}{2}\argsS{\vec{y} - \hat{f}(\vec{x}, \vec{w})}{^2},
\]
where $\vec{z} = (\vec{x}, \vec{y})$ projects to structure $Z = (X, Y)$ and $\vec{w}$ projects to $W$. The update rule is then of the form
\[
\vec{w}_{t+1} = \vec{w}_t - \eta_t \argsS{\vec{y}_t-\hat{f}(\vec{x}_t, \vec{w}_t)}{^{\T}}g(\vec{x}_t, \vec{w}_t),
\]
where $g(\vec{x}_t, \vec{w}_t) \in \partial \hat{L}_{mse}(\vec{z}_t, \vec{w}_t)$ is a stochastic generalized gradient of the lifted loss at $\vec{w}_t$.

\paragraph*{Structure Quantization.}
Structure quantization generalizes vector quantization to the quantization of structures. For graphs, a number of structure quantizer design techniques for the purpose of central clustering have already been proposed. Examples include competitive learning  \cite{Gold:1996b,Gunter:2002,Jain:2009b} and k-means as well as k-medoids algorithms \cite{Ferrer:2009b,Jain:2008,Schenker:2007}.

Let $\S{W} = \S{X}_\Gamma^k$ be the parameter space and let $\S{Z} = \S{X}_\Gamma $ be the space of observable data. The parameter space $\S{W}$ consists of $k$-tuples
$W = \args{W_1, \ldots, W_k}$, called \emph{codebook}.

The general loss function of structure quantization is defined by the distortion
\[
L_{sq}(X, W) = \min_{1 \leq i \leq k} d(X,W_i).
\]
For generalized differentiable distance function $d(\cdot|\cdot)$, the update rule is defined by
\[
\vec{w}_{t+1}^{*} = \vec{w}_t^*  - \eta g(\vec{x}_t, \vec{w}_t^*),
\]
where $(\vec{x}_t, \vec{w}_t^*)$ is an optimal alignment of input structure $X_t$ and its closest codebook structure $W_t^*$. If $d(\cdot|\cdot)$ is the squared intrinsic metric, we have
$g(\vec{x}, \vec{w}_t^*) = \vec{x}_t -\vec{w}_t^*$.

Observe that structure quantization also generalizes the problem of estimating a mean graph of Section 2.4 by fixing the number $k$ of centroids to $1$.

\section{Conclusion}
This contribution proves consistency of learning in structured domains by reducing it to stochastic generalized gradient learning on Riemannian orbifolds. The proposed framework is applicable to learning on combinatorial structures such as point patterns, trees, and graphs. In retrospect, the proposed results provide a theoretical foundation and statistical justification of a number of existing learning methods that directly operate in the domain of graphs. In addition, the orbifold framework provides a generic technique to generalize gradient-based learning methods to structured domains. Future work aims at generalizing the theory to more general Riemannian orbifolds and to discontinuous graph edit distance functions.

\section*{Acknowledgments.}
The authors are very grateful to Vladimir Norkin for his kind support and valuable comments.
\bibliographystyle{plain}
\bibliography{spr}

\end{document}